\documentclass[sigconf]{acmart}
\setcopyright{none}
\renewcommand\footnotetextcopyrightpermission[1]{}
\acmConference{}{}{}

\usepackage{booktabs}
\usepackage{multirow}
\usepackage{graphicx}
\usepackage{bm}
\usepackage{makecell}
\usepackage{pifont}
\usepackage{graphicx}
\usepackage{tabularx}
\usepackage{array}
\usepackage{graphicx}
\usepackage{subcaption}

\AtBeginDocument{%
  }

\begin{document}

\title{What Makes Graph Unified? Principles and Generative Sliding-Window Transformer for Graph Foundation Models}

\author{Dongxiao He}
\email{hedongxiao@tju.edu.cn}
\affiliation{
  \institution{School of Computer Science and Technology Tianjin University}
  \city{Tianjin}
  \country{China}
}

\author{Siqi Liu}
\email{siqiliu@tju.edu.cn}
\affiliation{
  \institution{School of Computer Science and Technology Tianjin University}
  \city{Tianjin}
  \country{China}
}

\author{Jitao Zhao}
\authornote{Corresponding author.}
\email{zjtao@tju.edu.cn}
\affiliation{
  \institution{School of Computer Science and Technology Tianjin University}
  \city{Tianjin}
  \country{China}
}

\author{Yawen Li}
\email{warmly0716@126.com}
\affiliation{
 \institution{School of Economics and Management Beijing University of Posts and Telecommunications}
 \city{Beijing}
 \country{China}
}

\author{Yi Wang}
\email{wangyi076@tju.edu.cn}
\affiliation{
  \institution{School of Computer Science and Technology Tianjin University}
  \city{Tianjin}
  \country{China}
}

\author{Di Jin}
\email{jindi@tju.edu.cn}
\affiliation{
  \institution{School of Computer Science and Technology Tianjin University}
  \city{Tianjin}
  \country{China}
  }

\renewcommand{\shortauthors}{Dongxiao He et al.}

\begin{abstract}
Graph Foundation Models (GFMs) have recently emerged as a promising paradigm for general-purpose graph learning, aiming to learn reusable knowledge that generalizes across diverse graph domains and downstream tasks, reducing the need for specific model development. 
Achieving this goal requires reconciling the substantial heterogeneity in node features, graph structures, and semantic information across domains. Among them, heterogeneous node features constitute a fundamental input-level barrier, as their dimensionality and semantics vary substantially across datasets. 
Existing studies typically project or map heterogeneous node features into a fixed-dimensional space, often implicitly equating dimensional uniformity with effective feature unification. Yet dimensional consistency alone does not ensure that the unified features preserve informative semantics and capture transferable patterns that can support cross-domain knowledge transfer.
To bridge this conceptual gap, we distill four desiderata for cross-domain graph feature unification: formal uniformity, cross-domain transferability, information preservation, and backbone compatibility.
Guided by these principles, we propose SliGFM, a graph foundation model built upon topology-aware sliding-window feature encoding and generative reconstruction. 
SliGFM orders feature dimensions by topological smoothness and scans the reordered features with a shared sliding-window feature encoder, transforming heterogeneous features into a common space of ordered fixed-dimensional feature tokens. 
This formulation enables a smoothness-aware transformer to capture transferable relational patterns among feature tokens within each node, while the generative reconstruction objective encourages preservation of the original feature information. 
Finally, SliGFM dynamically combines multi-hop representations by graph-level structural statistics, adapting neighborhood aggregation to different graph structures.
Extensive experiments demonstrate the effectiveness and transferability of SliGFM across diverse graph datasets.
\end{abstract}

\ccsdesc[500]{Networks~Network algorithms}
\ccsdesc[300]{Information systems~Data mining}
\ccsdesc[300]{Computing methodologies~Neural networks}

\keywords{Graph Neural Networks; Graph Foundation Models; Cross-domain Graph Learning}

\maketitle

\section{Introduction}
Recently, Graph Foundation Models (GFMs) have emerged as a new and promising paradigm for graph learning~\cite{GFM_survey1, GFM_survey2}. Drawing on the success of foundation models in computer vision and natural language processing~\cite{FM_LM, FM_CV}, GFMs pretrain on large-scale collections of diverse graphs to capture generalizable knowledge that can be transferred to previously unseen graph domains and downstream tasks~\cite{GraphAny, BRIDGE, OFA}. This paradigm departs from the conventional practice of developing and training a separate model for each dataset and task, offering a unified modeling framework that can accommodate structurally and semantically diverse graphs.

This pursuit is fundamentally constrained by the pronounced heterogeneity of graph data across domains. Graphs collected from different domains routinely differ in their feature spaces, topological patterns, and semantic information, making it difficult for a shared backbone to process them consistently~\cite{OFA, MDGFM, GIT, AnyGraph}. Among them, node features constitute a fundamental input-level obstacle. Unlike images or text, whose inputs are organized around canonical units like pixels or vocabularies, node features are defined independently by each dataset and vary in dimensionality, value distribution, and semantic meaning. As a result, the backbone has no principled way to interpret them consistently~\cite{TSGFM, GraphAlign, ALL_In}. Effective cross-domain feature unification is therefore a prerequisite for realizing the knowledge reuse goal of GFMs.

\begin{figure}[t]
\centering
\includegraphics[width=0.45\textwidth]{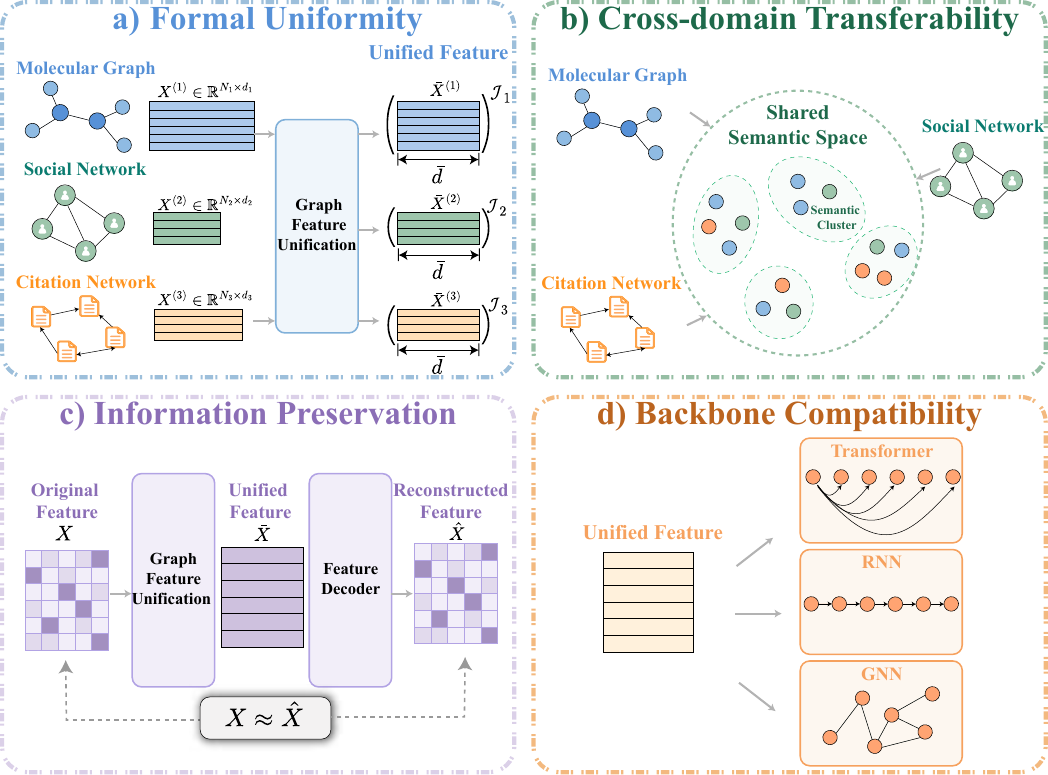} 
\caption{Schematic illustration of the four desiderata for cross-domain graph feature unification.}
\label{fig:desiderata}
\vspace{-0.6cm}
\end{figure}

To address this input-level incompatibility, existing studies have developed various mechanisms for feature transformation. A straightforward approach is to use statistical projections, such as principal component analysis (PCA) or singular value decomposition (SVD), to convert heterogeneous feature matrices into fixed-dimensional inputs~\cite{MDGPT, SAMGPT, MDGFM}. However, projection bases estimated independently for each domain may lack consistent coordinate semantics and discard informative content. More recent methods avoid coordinate-wise correspondence by deriving transferable representations from feature distributions, dependencies, or metadata~\cite{FUG, STAGE, TIG}. In parallel, language-mediated approaches textualize graphs and use pretrained language models to establish a shared semantic interface across domains~\cite{OFA, ZeroG, UniGraph}. Although these approaches map heterogeneous node features into a shared representation space, the existence of such a space alone does not ensure that the unified features preserve informative semantics or capture transferable patterns that support cross-domain knowledge transfer.

This limitation suggests that the effectiveness of cross-domain graph feature unification should be evaluated by the functional properties of unified results, rather than the construction mechanism solely. Accordingly, we formulate four desirable properties of effective feature unification, as illustrated in Figure~\ref{fig:desiderata}: (1) \textbf{Formal Uniformity}: heterogeneous features are mapped to a consistent feature specification across domains; (2) \textbf{Cross-domain Transferability}: unified features admit a domain-invariant semantic interpretation, allowing semantic patterns to be reused across domains; (3) \textbf{Information Preservation}: unified features retain sufficient information to recover their domain-specific inputs; and (4) \textbf{Backbone Compatibility}: unified features are organized in accordance with the input structure and inductive bias of the target backbone, enabling the backbone to effectively exploit them. These desiderata provide a principled basis for assessing whether a feature transformation achieves effective unification and facilitates knowledge reuse in GFMs. Formal definitions and detailed discussions are provided in Section~\ref{sec:3.1}.

Motivated by these desiderata, we propose SliGFM, a graph foundation model built around topology-aware sliding-window feature modeling and generative reconstruction. Rather than treating dimensional consistency as sufficient for feature unification, SliGFM exploits the variation of feature dimensions over topology as a reference that remains comparable across domains. Specifically, SliGFM first orders feature dimensions by topological smoothness, partitions the reordered features into overlapping windows, and maps them through a parameter-shared sliding-window encoder into fixed-dimensional tokens. This tokenization yields a formally unified input organization that is inherently compatible with transformer backbones. Within this shared organization, an intra-node transformer equipped with a relative-smoothness bias then models relations among the tokens, allowing dependency patterns at comparable variation scales to transfer across graphs. To prevent shared tokenization from erasing domain-specific information, a generative decoder reconstructs the original ordered feature patches during pretraining. Finally, SliGFM accommodates structural heterogeneity across graphs by adapting the scale of neighborhood aggregation to each graph, rather than imposing a fixed receptive field across domains.

In summary, our main contributions are:
\begin{itemize}
    \item We distill four desiderata for cross-domain graph feature unification: formal uniformity, cross-domain transferability, information preservation, and backbone compatibility.

    \item Guided by these desiderata, we propose SliGFM, a graph foundation model that unifies heterogeneous node features via topology-aware sliding-window tokenization coupled with a generative reconstruction objective, yielding transferable node-level representations across graph domains.

    \item We conduct extensive experiments on cross-domain node classification and graph classification, providing empirical evidence for the effectiveness and transferability of SliGFM.
    
\end{itemize}

\section{Preliminaries}

\paragraph{Graph.}
An attributed graph is denoted as $\mathcal{G}=(\mathcal{V},\mathcal{E},\mathbf{X})$, where $\mathcal{V}=\{v_1,\ldots,v_N\}$ is the set of $N$ nodes, $\mathcal{E}\subseteq\mathcal{V}\times\mathcal{V}$ is the set of edges, and $\mathbf{X}\in\mathbb{R}^{N\times d}$ is the node feature matrix with $d$-dimensional node feature. The graph topology can be represented by an adjacency matrix $\mathbf{A}\in\{0,1\}^{N\times N}$, where $\mathbf{A}_{ij}=1$ if $(v_i,v_j)\in\mathcal{E}$ and $\mathbf{A}_{ij}=0$ otherwise. For a node $v_i$, we use $\mathcal{N}(v_i)=\{v_j\mid (v_i,v_j)\in\mathcal{E}\}$ to denote its one-hop neighborhood. The notations used in this paper and their descriptions are summarized in Appendix~\ref{appendix:notations}.

\paragraph{Cross-Domain Graph Feature Unification.}
We consider a collection of graphs $\mathcal{D}_{\mathcal{G}}=\{\mathcal{G}^{(1)},\mathcal{G}^{(2)},\ldots, \mathcal{G}^{(M)}\}$ from different graph domains. The $m$-th graph is denoted as $\mathcal{G}^{(m)}=(\mathcal{V}^{(m)},\mathcal{E}^{(m)},\mathbf{X}^{(m)})$, where $|\mathcal{V}^{(m)}|=N_m$ and $\mathbf{X}^{(m)}\in\mathbb{R}^{N_m\times d_m}$. Across domains, node feature may have different dimensions $d_m$ and semantics, which is referred to as feature heterogeneity. To mitigate feature heterogeneity and obtain a consistent input representation, we define a cross-domain graph feature unification mechanism $\mathcal{U}$, which maps domain-specific node feature into a shared feature space: 
\begin{equation}
    \bar{\mathbf{X}}^{(m)} = \mathcal{U}(\mathbf{X}^{(m)}) \in \bar{\mathcal{X}},
     \qquad
    \forall m\in\{1,\ldots,M\},
\end{equation}
where $\bar{\mathcal{X}}$ denotes the common representation space shared by all graph domains. Consequently, although the original feature spaces may vary across domains, the resulting representations follow a common representation specification and can be consistently processed by a shared graph foundation model backbone.

\paragraph{Graph Foundation Models.}
By pretraining on large-scale diverse graph data, graph foundation models (GFMs) seek to learn generalizable, transferable representations applicable across different datasets, tasks, and application scenarios. The pretrained model can then be adapted to specific downstream tasks through fine-tuning or prompt-based learning, particularly under limited supervision.

\section{Cross-Domain Graph Feature Unification}
\label{sec:3}
An effective cross-domain feature unification mechanism should reconcile heterogeneous feature spaces while preserving transferable information and remaining compatible with diverse graph backbones. In this section, we first establish four desirable properties for such a mechanism and then assess existing approaches.

\subsection{Four Desirable Properties}
As illustrated in Figure~\ref{fig:desiderata}, an effective cross-domain graph feature unification mechanism should satisfy four complementary properties, covering formal uniformity, cross-domain transferability, information preservation, and backbone compatibility.

\label{sec:3.1}
\paragraph{Formal Uniformity.}
The most fundamental requirement for cross-domain graph feature unification is formal uniformity, aiming to make originally heterogeneous features computationally compatible with a shared backbone. Specifically, the unified representations should follow a common representation specification, including a consistent numerical value space, a domain-independent feature dimension $\bar d$, and a common indexing convention. Formally, the unified representation satisfies:
\begin{equation}
    \bar{\mathbf X}^{(m)}
    = \mathcal U\!\left(\mathbf X^{(m)}\right)
    = \left(\bar{\mathbf x}^{(m)}_i\right)_{i\in\mathcal I_m},
    \qquad
    \bar{\mathbf x}^{(m)}_i\in\mathbb R^{\bar d},
\end{equation}
where $\mathcal I_m$ denotes the index set of the unified feature units associated with $\mathcal G^{(m)}$, so that every indexed unit is represented by a feature vector of the same dimension $\bar d$. For node-wise unification, $\mathcal I_m=\mathcal V^{(m)}$, in which case the unified representation can be equivalently organized as $\bar{\mathbf X}^{(m)}\in\mathbb R^{N_m\times\bar d}$.

\paragraph{Cross-Domain Transferability.}
Formal uniformity only ensures that a shared backbone can process heterogeneous graph features; cross-domain transferability further requires the unified representations to admit a domain-invariant semantic interpretation. For each unified feature unit $\bar{\mathbf{x}}_i^{(m)}$, let $s_i^{(m)}\in\mathcal{S}$ denote its associated semantics, where $\mathcal{S}$ is a semantic space shared across all graph domains. Cross-domain transferability requires the existence of a semantic mapping $\rho:\mathbb{R}^{\bar d}\rightarrow\mathcal{S}$, common to all domains, such that:
\begin{equation}
    d_{\mathcal{S}}\!\left(
        \rho\!\left(\bar{\mathbf{x}}_i^{(m)}\right),
        s_i^{(m)}
    \right)
    \leq \epsilon_{\mathrm{tr}},
    \qquad
    \forall m\in\{1,\ldots,M\},\quad
    \forall i\in\mathcal{I}_m,
\end{equation}
where $d_{\mathcal{S}}$ is a metric on $\mathcal{S}$ and $\epsilon_{\mathrm{tr}}$ denotes the tolerated semantic discrepancy. Concretely, consider any two graphs $m$ and $n$, and two feature units $i\in\mathcal{I}_m$ and $j\in\mathcal{I}_n$ with identical semantics, i.e., $s_i^{(m)}=s_j^{(n)}$. Since both feature units satisfy the preceding condition under the shared mapping $\rho$, the triangle inequality gives:
\begin{equation}
    d_{\mathcal{S}}\!\left(
        \rho\!\left(\bar{\mathbf{x}}_i^{(m)}\right),
        \rho\!\left(\bar{\mathbf{x}}_j^{(n)}\right)
    \right)
    \leq 2\epsilon_{\mathrm{tr}}.
\end{equation}
Thus, the semantic interpretation of a unified feature remains stable across domains, enabling semantic patterns learned in one graph domain to be reused in others.

\paragraph{Information Preservation.}
Feature unification should achieve cross-domain compatibility without irreversibly discarding the information contained in the original node features. Although the unified representations follow a common representation specification, they should retain sufficient information to reconstruct their domain-specific inputs. Let $\mathcal{X}_m$ denote the admissible feature space of domain $m$. Ideally, there should exist a domain-specific decoder $\operatorname{Dec}^{(m)}$ such that:
\begin{equation}
    \operatorname{Dec}^{(m)}
    \!\left(
        \mathcal{U}(\mathbf{X})
    \right)
    =
    \mathbf{X}^{(m)},
    \qquad
    \forall \mathbf{X}^{(m)}\in\mathcal{X}_m,\ m\in\{1,\ldots,M\}.
\end{equation}
Equivalently, the composition $\operatorname{Dec}^{(m)}\circ\mathcal{U}$ acts as the identity mapping on $\mathcal{X}_m$. This condition requires $\mathcal{U}$ to be injective over each domain-specific feature space, preventing distinct original features from being irreversibly collapsed into the same representation.

In practice, exact reconstruction may be unattainable. Let $P_m$ denote the distribution of admissible feature matrices in domain $m$, and let $\ell_m$ be an appropriate domain-specific reconstruction loss. Information preservation can then be relaxed by requiring:
\begin{equation}
    \mathbb{E}_{\mathbf{X}\sim P_m}
    \left[
        \ell_m\!\left(
            \mathbf{X},
            \operatorname{Dec}^{(m)}
            \!\left(\mathcal{U}(\mathbf{X})\right)
        \right)
    \right]
    \leq
    \epsilon_{\mathrm{rec}},
    \qquad
    \forall m\in\{1,\ldots,M\},
\end{equation}
where $\epsilon_{\mathrm{rec}}\geq 0$ denotes the tolerated reconstruction distortion. For real-valued node features, $\ell_m$ can be instantiated as the squared Frobenius reconstruction error. The ideal lossless case corresponds to $\epsilon_{\mathrm{rec}}=0$, while a small positive value allows approximate information preservation in practical implementations.

\paragraph{Backbone Compatibility.}
Beyond satisfying a common numerical specification, unified representations should conform to the input structure and inductive bias of the target backbone, so that the backbone's computation can effectively exploit their organization rather than merely accept them as formally valid input. For instance, RNNs require consistent cross-domain feature ordering; Transformers rely on appropriate positional or structural encodings; and for low-pass message-passing GNNs, unified node features must further exhibit topology-consistent local smoothness, so that neighborhood aggregation reinforces rather than distorts compatible signals.

\subsection{Assessment of Existing Approaches}
\label{sec:assessment}
Existing approaches to cross-domain graph feature unification can be roughly divided into statistical subspace projection, transfer-invariant encoding, and language-mediated alignment. 

Statistical subspace projection first uses statistical decomposition to align feature dimensions. MDGPT and SAMGPT apply singular value decomposition (SVD), whereas MDGFM uses principal component analysis (PCA) \cite{MDGPT, SAMGPT, MDGFM}. After dimensional alignment, MDGPT introduces domain tokens to calibrate domain-specific semantics, SAMGPT employs structure tokens to coordinate aggregation patterns across domains, and MDGFM combines domain and shared tokens with structure learning for further semantic and topological alignment. Although these pipelines establish formal uniformity, their projection bases are estimated from domain-specific feature statistics, so the aligned coordinates do not inherently carry consistent semantics across domains. Moreover, low-rank projection discards components outside the retained subspace, potentially causing irreversible information loss.

Unlike independently estimated statistical projections, transfer-invariant encoding extracts shared cross-domain references from feature distributions, dependencies, or metadata. FUG models the distribution of each feature dimension to generate a learnable basis transformation matrix, jointly optimized with the graph encoder via relative-relation and global-uniformity objectives~\cite{FUG}. STAGE builds an edge-level graph from the marginal and conditional probabilities of endpoint attributes, obtaining fixed-dimensional edge representations through intra-edge encoding and inter-edge propagation~\cite{STAGE}. TIG extracts transfer-invariant metadata to unify heterogeneous features into structural representations~\cite{TIG}. While these methods largely achieve formal uniformity and cross-domain transferability, they generally do not offer an explicit mechanism to reconstruct or otherwise guarantee preservation of the original input information. Moreover, the resulting feature organization is not tailored to the input conventions and inductive biases of the backbones, leaving practical compatibility unverified.

Another line of work leverages natural language as a shared semantic interface, using pretrained language models or large language models (LLMs) to align heterogeneous graphs. Methods such as OFA and ZeroG express node and edge features and class semantics in text and encode them with a shared language model~\cite{OFA, ZeroG}. UniGraph instead couples a language encoder with a graph encoder, allowing textual semantics and graph structure to be modeled jointly~\cite{UniGraph}. LLaGA further organizes graph neighborhoods using structural templates and maps the resulting graph tokens into a pretrained LLM's embedding space, enabling the LLM to serve as the reasoning backbone for graph tasks~\cite{LLaGA}. Although these methods exploit the powerful representation capacity of LLMs to provide fixed-dimensional inputs that satisfy formal uniformity and establish a shared semantic reference across domains, they rely on graphs being faithfully expressible in natural language, which limits their applicability to purely numerical or otherwise not readily amenable to textualization. Moreover, textualizing graphs may discard fine-grained information from the original inputs, offering no guarantee that the original features can be exactly recovered from the resulting text.

\begin{figure*}[t]
\centering
\includegraphics[width=1\textwidth]{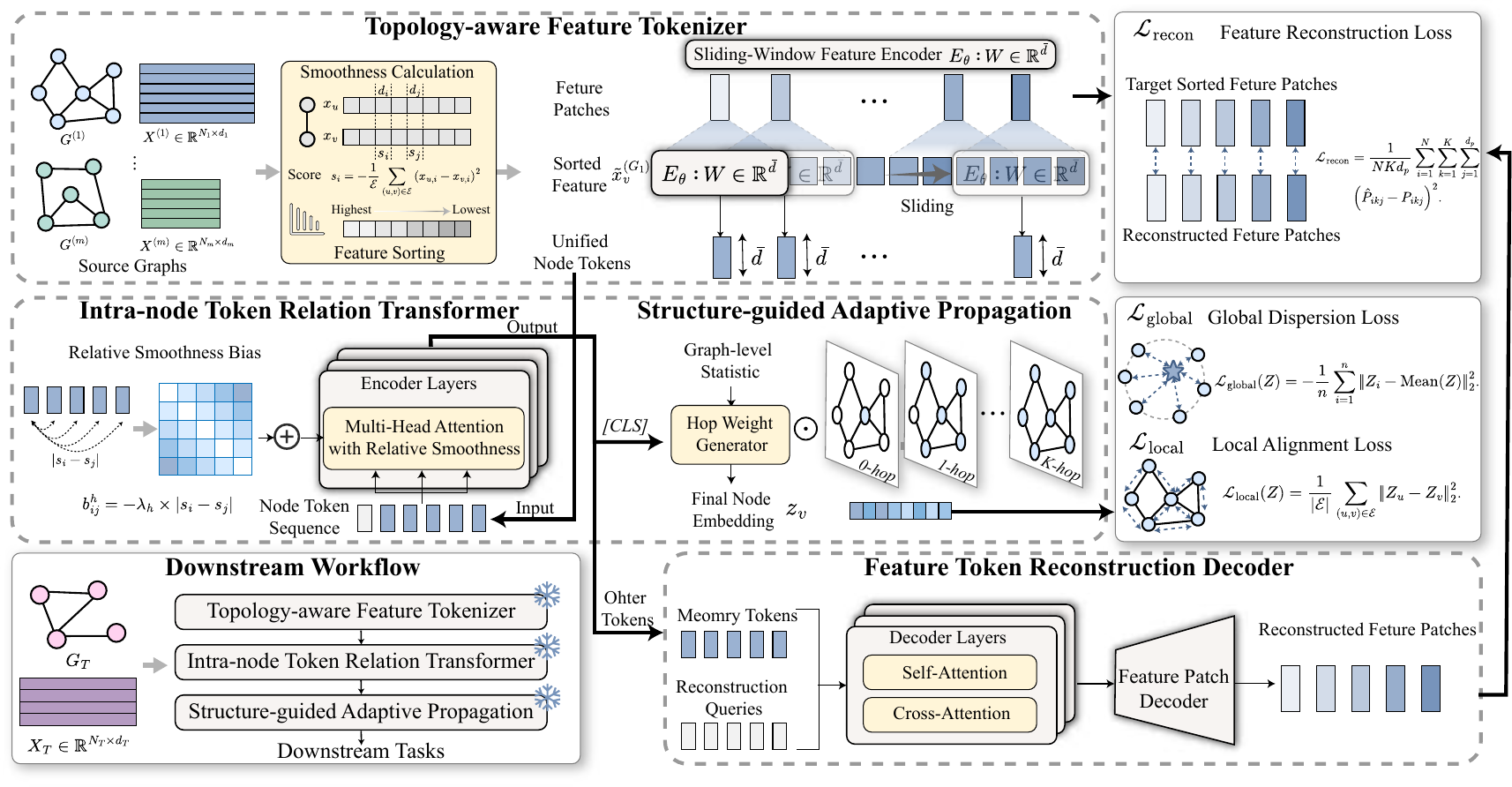} 
\caption{The overview of the proposed SliGFM model. }
\label{model}
\vspace{-0.3cm}
\end{figure*}

\section{Methodology}

\subsection{Overview}
As illustrated in Figure~\ref{model}, our model consists of four core components for learning transferable representations from cross-domain graphs with heterogeneous features and structures.
Specifically, the Topology-aware Feature Tokenizer first sorts node features by the topological smoothness of each feature dimension and employs a parameter-shared sliding-window encoder to map local feature patches into a fixed-dimensional token sequence. 
Subsequently, a node-specific \textit{[CLS]} token is prepended to this sequence, and the Intra-node Token Relation Transformer equipped with a relative smoothness bias captures dependencies among feature subspaces of varying smoothness. 
The resulting \textit{[CLS]} representation that aggregates the complete node feature information is then fed into the Structure-guided Adaptive Propagation module, which aggregates multi-hop neighborhood representations and adaptively fuses them using weights conditioned on graph statistics, thereby accommodating structural heterogeneity across different graph domains and producing the final node representation.
In parallel, a Feature Token Reconstruction branch uses learnable queries to recover the original feature patches from the contextualized tokens, encouraging the encoder to preserve complete attribute information.
The overall model is jointly optimized with a reconstruction loss that supervises the recovery of original feature patches, together with a global dispersion constraint and a local alignment constraint applied to the final representation to avoid representation collapse and preserve structural consistency among neighboring nodes. 

\subsection{Topology-aware Feature Tokenization}
Node features across graph datasets vary in dimensionality and semantics, and their original order typically offers no consistent regular pattern for a shared model to exploit. Mapping them directly along this order may cause the same position to encode entirely different features across graphs, limiting model transferability. To address this issue, we introduce a topology-aware feature tokenizer that establishes a relatively consistent feature order according to smoothness and maps node features of varying dimensionality into a unified token space.

For the $m$-th graph $\mathcal{G}^{(m)}=(\mathcal{V}^{(m)},\mathcal{E}^{(m)},\mathbf{X}^{(m)})$,
with $\mathbf{X}^{(m)}\in\mathbb{R}^{N_m\times d_m}$, we quantify the smoothness of each feature dimension by the average feature variation between adjacent nodes:
\begin{equation}
s_j^{(m)}
=
-\frac{1}{\left|\mathcal{E}^{(m)}\right|}
\sum_{(u,v)\in\mathcal{E}^{(m)}}
\left(x_{u,j}^{(m)}-x_{v,j}^{(m)}\right)^2 ,
\label{eq:feature_smoothness}
\end{equation}
where $\mathbf{x}^{(m)}_{:,j}$ denotes the $j$-th feature dimension. We then sort all feature dimensions in descending order of smoothness, yielding the permutation
$\boldsymbol{\pi}^{(m)}
=\operatorname{argsort}_{\downarrow}
(s_1^{(m)},\ldots,s_{d_m}^{(m)})$, and the  reordered feature matrix $\bar{\mathbf{X}}^{(m)}
=\mathbf{X}^{(m)}_{:,\boldsymbol{\pi}^{(m)}}$.
By ordering features according to their variation over the graph topology, this procedure reduces dependence on dataset-specific column order and provides an ordered input for the subsequent smoothness-aware token modeling. After reordering, each node feature $\bar{\mathbf{x}}_i^{(m)}$ is segmented along the feature dimension using a sliding window of size $d_p$ and stride $\delta$. The $k$-th local feature patch of $\bar{\mathbf{x}}_i^{(m)}$, starting at index $a_k=(k-1)\delta+1$, is thus givn by
$\mathbf{p}_{i,k}^{(m)}
=
\left[
\bar{x}_{i,a_k}^{(m)},
\bar{x}_{i,a_k+1}^{(m)},
\ldots,
\bar{x}_{i,a_k+d_p-1}^{(m)}
\right]
\in\mathbb{R}^{d_p}$.
Notice that if the final window extends beyond the original feature dimension, its missing entries are zero-padded. 

Rather than training a separate encoder for each dataset's distinct dimensionality, we design a single, parameter-shared sliding-window encoder that maps feature patches from all source graphs into a common space. Specifically, we introduce $E_{\theta}: \mathbb{R}^{d_p}\to\mathbb{R}^{d_t}$ maps each patch to a fixed-dimensional token as:
\begin{equation}
\begin{gathered}
\mathbf{t}_{i,k}^{(m)}
=
E_{\theta}\!\left(\mathbf{p}_{i,k}^{(m)}\right)
\in\mathbb{R}^{d_t},
\qquad
\\
\mathbf{T}_{i}^{(m)}
=
\left[
\mathbf{t}_{i,1}^{(m)};
\mathbf{t}_{i,2}^{(m)};
\ldots;
\mathbf{t}_{i,K_m}^{(m)}
\right]
\in\mathbb{R}^{K_m\times d_t},
\end{gathered}
\end{equation}
where $d_t$ is the unified token dimension, and $K_m$ is the number of tokens. The tokenizer not only unifies the input format but also the structural semantics of heterogeneous features, as features with similar smoothness are assigned to nearby positions and grouped into local tokens representing comparable scales of topological variation. Consequently, tokens from different graphs reflect comparable topology-induced functional semantics, allowing the shared encoder to reuse consistent topology–attribute patterns.

\subsection{Intra-node Token Relation Transformer}
The topology-aware feature tokenizer represents each node as a sequence of local feature tokens. To integrate these tokens into a holistic node representation, we introduce an intra-node token relation transformer to model their interactions. For each node $v_i^{(m)}$, we first prepend a node-specific $\textit{[CLS]}$ token $\mathbf{c}_i^{(m)}\in\mathbb{R}^{d_t}$ to its feature-token sequence $\mathbf{T}_i^{(m)}$ as
$\mathbf{U}_i^{(m,0)}
=
[\mathbf{c}_i^{(m)};
\mathbf{t}_{i,1}^{(m)};
\ldots;
\mathbf{t}_{i,K_m}^{(m)}]
\in\mathbb{R}^{(K_m+1)\times d_t}$. The $\textit{[CLS]}$ token is initialized using the PCA-projected topology-ordered features, providing an initial global summary of the node features. 

We then feed $\mathbf{U}_i^{(m,0)}$ into a standard $L$-layer Transformer encoder~\cite{Transformer} with a relative smoothness bias $\mathbf{B}^{(m)}$. Following ALiBi~\cite{ALiBi}, we introduce a head-specific linear bias into the attention logits, replacing the original positional distance with the relative smoothness between feature tokens:
\begin{equation}
\mathbf{H}
=
\operatorname{softmax}
\left(
\frac{\mathbf{Q}\mathbf{K}^{\top}}{\sqrt{d_h}}
+
\mathbf{B}^{(m)}
\right)\mathbf{V}.
\end{equation}
To construct the bias, we first define the token-level smoothness $\bar{s}_k^{(m)}$ of the $k$-th feature token by averaging the feature-level smoothness scores within the corresponding patch, excluding padded positions. The relative smoothness bias of the $h$-th head between two feature tokens is then defined as:
\begin{equation}
B_{k\ell}^{(h,m)}
=
\begin{cases}
-\lambda_h
\left|
\bar{s}_k^{(m)}-\bar{s}_\ell^{(m)}
\right|,
& k,\ell\in\{1,\ldots,K_m\},\\[3pt]
0,
& k=0\ \text{or}\ \ell=0.
\end{cases}
\end{equation}
where $\lambda_h>0$ is a head-specific scaling coefficient. Such bias acts as a topology-aware prior by penalizing token pairs in proportion to their smoothness discrepancy. Consequently, self-attention remains content-dependent while favoring interactions between feature subspaces with similar topology-induced variation patterns.

After $L$ Transformer layers, we obtain the contextualized sequence
$\mathbf{U}_i^{(m,L)}
=
[\mathbf{c}_{i}^{(m)};
\widehat{\mathbf{t}}_{i,1}^{(m)};
\ldots;
\widehat{\mathbf{t}}_{i,K_m}^{(m)}]$.
The $\textit{[CLS]}$ output $\mathbf{c}_{i}^{(m)}$ is taken as the representation of node $v_i^{(m)}$. This Transformer encoder captures intra-node attribute dependencies, while inter-node structural interactions are modeled by the subsequent propagation.

\subsection{Structure-guided Adaptive Propagation}

The preceding Transformer module models feature interactions within each node but does not capture inter-node dependencies. We therefore propagate its outputs over multiple hops to incorporate neighborhood context. Since graphs vary substantially in sparsity, connectivity, and local clustering, a fixed propagation range may not be suitable for all graphs. 

To account for such structural heterogeneity, we characterize each graph using its degree histogram, clustering-coefficient histogram, and log-scaled motif counts, including triangles and short cycles~\cite{Graph_Statistic}. We then employ an MLP as the hop-weight generator to map these statistics to graph-specific propagation weights:
\begin{equation}
\boldsymbol{\alpha}^{(m)}
=
\operatorname{Softmax}
\left(
\operatorname{MLP}(\mathbf{g}^{(m)})
\right),
\end{equation}
where $\mathbf{g}^{(m)}$ denotes the $m$-th graph statistics, and $\boldsymbol{\alpha}^{(m)}$ determines the contribution of the $\ell$-hop representation. 

Using the graph-specific weights $\boldsymbol{\alpha}^{(m)}$, we propagate the node representations $\mathbf{C}^{(m)}$ over different hop ranges and combine the resulting multi-scale representations as:
\begin{equation}
\mathbf{Z}^{(m)}
=
\left(
\sum_{\ell=0}^{K_{\mathrm{hop}}}
\boldsymbol{\alpha}^{(m)}
\left(\widehat{\mathbf{A}}^{(m)}\right)^{\ell}
\mathbf{C}^{(m)}
\right)
\mathbf{W}_{\mathrm{SGC}},
\label{eq:adaptive_sgc}
\end{equation}
where $\mathbf{W}_{\mathrm{SGC}}$ is a learnable transformation matrix. By conditioning the hop fusion on graph-level structural statistics, the module adaptively balances node features with neighborhood context. This avoids imposing a uniform receptive field across structurally heterogeneous graphs and alleviates insufficient propagation.

\subsection{Generative Feature Token Reconstruction}

While the preceding propagation module enriches the node representation with inter-node structural information, the contextualized feature tokens $\widehat{\mathbf{T}}_i^{(m)}$ retain fine-grained feature information. However, mapping heterogeneous features into a shared token space may obscure dataset-specific details during representation learning. We therefore introduce a generative feature-token reconstruction branch that recovers the original feature patches from the contextualized tokens, encouraging the shared encoder to preserve meaningful information while learning transferable representations.

Specifically, we treat
$\widehat{\mathbf{T}}_i^{(m)}
=
[\widehat{\mathbf{t}}_{i,1}^{(m)};
\ldots;
\widehat{\mathbf{t}}_{i,K_m}^{(m)}]$
as the memory of a Transformer decoder. To accommodate variable-length token sequences, we maintain a shared bank of learnable reconstruction queries and select the first $K_m$ queries for graph $\mathcal{G}^{(m)}$. The decoder generates one representation for each feature-token position:
\begin{equation}
\mathbf{O}_i^{(m)}
=
\operatorname{Decoder}_{\omega}
\left(
\mathbf{R}_{\mathrm{qry}}^{(K_m)},
\widehat{\mathbf{T}}_i^{(m)}
\right)
=
[\mathbf{o}_{i,1}^{(m)};
\ldots;
\mathbf{o}_{i,K_m}^{(m)}]
\in\mathbb{R}^{K_m\times d_t}.
\label{eq:reconstruction_decoder}
\end{equation}
Through self-attention among the queries and cross-attention to the contextualized tokens, the decoder jointly captures dependencies among reconstruction positions and retrieves the attribute information required for each patch.

Finally, a patch decoder $D_{\xi}:\mathbb{R}^{d_t}\rightarrow\mathbb{R}^{d_p}$, shared across all graphs, maps each decoder output back to the original patch space:
\begin{equation}
\widehat{\mathbf{p}}_{i,k}^{(m)}
=
D_{\xi}
\left(
\mathbf{o}_{i,k}^{(m)}
\right)
\in\mathbb{R}^{d_p},
\qquad
k=1,\ldots,K_m,
\label{eq:feature_patch_reconstruction}
\end{equation}
where $D_{\xi}$ is instantiated as an MLP. This reconstruction branch complements structure-guided propagation by preserving fine-grained feature content in the shared token representations.

\subsection{Training Objectives}

The overall training objective consists of three loss terms:

\paragraph{Feature reconstruction.}
We minimize the mean-squared error between the reconstructed feature patches
$\widehat{\mathbf{p}}_{i,k}^{(m)}$ and their original feature patches
$\mathbf{p}_{i,k}^{(m)}$ to constrain the shared encoder to preserve more complete node attribute semantics. 
\begin{equation}
\mathcal{L}_{\mathrm{recon}}
=
\frac{1}{M}
\sum_{m=1}^{M}
\frac{1}{N_m K_m d_p}
\sum_{i=1}^{N_m}
\sum_{k=1}^{K_m}
\left\|
\widehat{\mathbf{p}}_{i,k}^{(m)}
-
\mathbf{p}_{i,k}^{(m)}
\right\|_2^2.
\label{eq:reconstruction_loss}
\end{equation}

\paragraph{Local alignment.}
Let $\mathbf{z}_i^{(m)}$ denote the normalized final representation of node $v_i^{(m)}$. The local alignment loss minimizes the average distance between connected nodes to preserve local consistency:
\begin{equation}
\mathcal{L}_{\mathrm{local}}
=
\frac{1}{M}
\sum_{m=1}^{M}
\frac{1}{|\mathcal{E}^{(m)}|}
\sum_{(u,v)\in\mathcal{E}^{(m)}}
\left\|
\mathbf{z}_u^{(m)}
-
\mathbf{z}_v^{(m)}
\right\|_2.
\label{eq:local_alignment_loss}
\end{equation}

\paragraph{Global dispersion.}
To prevent excessive local alignment from collapsing the representation space, we employ the center-away scattering mechanism proposed in SGRL~\cite{SGRL} as our global objective:
\begin{equation}
\mathcal{L}_{\mathrm{global}}
=
-\frac{1}{M}
\sum_{m=1}^{M}
\frac{1}{N_m}
\sum_{i=1}^{N_m}
\left\|
\mathbf{z}_i^{(m)}
-
\bar{\mathbf{z}}^{(m)}
\right\|_2 ,
\quad
\bar{\mathbf{z}}^{(m)}
=
\frac{1}{N_m}
\sum_{i=1}^{N_m}
\mathbf{z}_i^{(m)}.
\label{eq:global_dispersion_loss}
\end{equation}

Finally, the overall objective is:
\begin{equation}
\mathcal{L}
=
\lambda_{\mathrm{recon}}\mathcal{L}_{\mathrm{recon}}
+
\lambda_{\mathrm{local}}\mathcal{L}_{\mathrm{local}}
+
\lambda_{\mathrm{global}}\mathcal{L}_{\mathrm{global}},
\label{eq:overall_objective}
\end{equation}
where the three coefficients balance attribute reconstruction, local alignment, and global dispersion.

\subsection{Tuning-free Downstream Transfer}

For an unseen target graph, SliGFM directly employs the pretrained backbone without any tuning techniques. The topology-aware tokenizer first converts its heterogeneous node features into unified tokens, which are subsequently integrated by the intra-node Transformer to obtain node-level representations. These representations are further enriched with target-specific structural information through structure-guided adaptive multi-hop propagation. During inference, the reconstruction decoder is discarded, and the resulting node embeddings are directly used for downstream prediction.

\section{Experiments}

\subsection{Experiment Setup}

\paragraph{Datasets.}
For cross-domain node classification, we use nine benchmark datasets including six homophilic datasets: Cora~\cite{Cora}, CiteSeer~\cite{CiteSeer}, PubMed~\cite{PubMed}, Photo~\cite{Photo}, Computers~\cite{Computers},  CS~\cite{CS}, and two heterophilic datasets: Chameleon~\cite{Chameleon_Squirrel}, Squirrel~\cite{Chameleon_Squirrel}, Actor~\cite{Actor}. For cross-domain graph classification, we use IMDB-BINARY~\cite{IMDB-BINARY}, ENZYMES~\cite{ENZYMES}, and DD~\cite{DD}. More detailed information on these datasets can be found in Appendix \ref{appendix:datasets}.

\paragraph{Baselines.}
For node classification, we compare our SliGFM with representative graph foundation models, including MDGFM~\cite{MDGFM}, MDGPT~\cite{MDGPT}, BRIDGE~\cite{BRIDGE},  FUG~\cite{FUG}, LEDA~\cite{LEDA}, SAMGPT~\cite{SAMGPT}, TIG~\cite{TIG}, and TFSGFM~\cite{TFSGFM}. For graph classification, we compare against representative graph pre-training including GPPT~\cite{GPPT}, GPF~\cite{GPF}, and GPF-plus~\cite{GPF}, and prompt-based methods including ULTRA~\cite{ULTRA}, SCORE~\cite{SCORE}, and TFSGFM~\cite{TFSGFM}. We additionally report the supervised GCN~\cite{GCN} as a reference for graph classification.

\paragraph{Implementation Details.}
For cross-domain node classification, we first jointly pre-train our model on Cora, PubMed, Computers, and Chameleon. After pre-training, the model is kept frozen and directly evaluated on few-shot node classification tasks across all benchmark datasets, without any task-specific fine-tuning or parameter updates. For each dataset, we randomly generate 500 few-shot tasks and report the average Accuracy over these 500 trials. For cross-domain graph classification, we follow the experimental settings of TFSGFM~\cite{TFSGFM} to conduct zero-shot graph classification experiments, and report the average classification Accuracy and Macro-F1 score as the evaluation metrics. More detailed experimental configurations are provided in the Appendix~\ref{appendix:experiment}.

\begin{table*}[t]
\centering
\caption{One-shot cross-domain node classification results (Accuracy \% $\pm$ std). The best results are highlighted in bold, and the second-best are underlined.}
\label{tab:one_shot_node_classification}
\setlength{\tabcolsep}{4pt}
\renewcommand{\arraystretch}{1.15}
\resizebox{\textwidth}{!}{
\begin{tabular}{lccccccccc}
\toprule
\textbf{Method}
& \textbf{Cora}
& \textbf{CiteSeer}
& \textbf{PubMed}
& \textbf{Computers}
& \textbf{Photo}
& \textbf{CS}
& \textbf{Squirrel}
& \textbf{Chameleon}
& \textbf{Actor} \\
\midrule

\textbf{MDGFM}
& $36.22 \pm 6.96$
& $35.07 \pm 6.94$
& $46.51 \pm 7.65$
& $52.51 \pm 10.21$
& $57.51 \pm 8.83$
& $69.86 \pm 6.39$
& $21.93 \pm 2.46$
& $24.66 \pm 3.83$
& $19.65 \pm 2.70$ \\

\textbf{MDGPT}
& $41.58 \pm 8.00$
& $40.47 \pm 8.23$
& $50.57 \pm 10.16$
& $53.47 \pm 10.42$
& $66.66 \pm 9.10$
& $66.84 \pm 7.73$
& $23.70 \pm 3.56$
& $24.99 \pm 3.83$
& $19.87 \pm 2.46$ \\

\textbf{BRIDGE}
& $42.66 \pm 8.16$
& $43.57 \pm 8.64$
& \underline{$53.95 \pm 8.86$}
& $54.90 \pm 9.42$
& $63.19 \pm 9.17$
& $70.40 \pm 7.81$
& $\bm{24.22 \pm 3.89}$
& $25.15 \pm 4.26$
& $19.88 \pm 1.82$ \\

\textbf{FUG}
& $45.76 \pm 9.54$
& $34.08 \pm 8.00$
& $52.50 \pm 10.47$
& $45.95 \pm 10.22$
& $56.88 \pm 8.45$
& $75.62 \pm 7.31$
& $20.68 \pm 1.16$
& $22.15 \pm 2.97$
& $20.54 \pm 2.88$ \\

\textbf{LEDA}
& $48.77 \pm 9.97$
& $43.34 \pm 10.61$
& $52.36 \pm 11.92$
& $53.48 \pm 11.54$
& $67.45 \pm 7.56$
& $69.28 \pm 9.23$
& $21.15 \pm 1.75$
& $25.25 \pm 3.35$
& $19.57 \pm 3.13$ \\

\textbf{SAMGPT}
& $44.34 \pm 7.33$
& $35.77 \pm 6.43$
& $46.02 \pm 8.74$
& $44.85 \pm 8.05$
& $56.70 \pm 7.99$
& $69.85 \pm 6.40$
& \underline{$23.75 \pm 3.82$}
& $24.92 \pm 3.87$
& $20.05 \pm 1.96$ \\

\textbf{TIG}
& $49.86 \pm 10.91$
& $42.14 \pm 6.94$
& $50.73 \pm 10.28$
& $50.93 \pm 11.05$
& $61.29 \pm 8.72$
& $74.81 \pm 6.98$
& $20.33 \pm 0.78$
& $24.38 \pm 4.52$
& $19.48 \pm 2.28$ \\

\textbf{TFSGFM}
& \underline{$53.11 \pm 8.84$}
& \underline{$47.24 \pm 9.50$}
& $53.66 \pm 10.33$
& $\underline{57.06 \pm 12.12}$
& \underline{$68.12 \pm 10.30$}
& \underline{$78.00 \pm 8.25$}
& $20.76 \pm 1.43$
& \underline{$25.33 \pm 3.98$}
& \underline{$22.97 \pm 3.44$}\\
\midrule

\textbf{SliGFM(Ours)}
& $\bm{55.73 \pm 8.36}$
& $\bm{47.32 \pm 9.25}$
& $\bm{57.77 \pm 8.36}$
& $\bm{57.45\pm11.69}$
& $\bm{68.48\pm 9.04}$
& $\bm{79.45 \pm 6.29}$
& $21.21\pm2.17$
& $\bm{29.24 \pm 3.35}$
& $\bm{23.73\pm3.84}$ \\

\bottomrule
\end{tabular}
}
\end{table*}

\subsection{Few-shot Node Classification}
Table~\ref{tab:one_shot_node_classification} reports the one-shot node classification results across nine graph datasets. SliGFM achieves the best performance on eight datasets, attaining an average accuracy of 48.93\%, surpassing the strongest baseline TFSGFM by 1.57 percentage points. The improvements are particularly pronounced on Chameleon, PubMed, and Cora, where SliGFM exceeds the best competing results by 3.91, 3.82, and 2.62 percentage points, respectively. 

We attribute these gains primarily to the way SliGFM handles heterogeneous node features and topology structure. Existing methods often rely on dataset-specific feature coordinates or compress heterogeneous feature into a shared space, which may weaken the knowledge transferred across graphs with different feature dimensions and semantics. In contrast, SliGFM reorganizes feature dimensions according to their topology-aware smoothness and models local relations within the resulting feature-token sequence, providing a more consistent basis for cross-domain transfer. Meanwhile, the reconstruction objective encourages the unified representations to retain informative content from the original feature. Such information preservation is particularly important under few-shot supervision, where only a few labeled examples are available to compensate for information lost during feature transformation.  Moreover, the adaptive combination of different propagation ranges reduces the dependence on source-specific structural patterns, enabling the pretrained model to better accommodate diverse target-graph topologies without downstream tuning. Additional results under other few-shot settings are provided in Appendix~\ref{appendix:few-shot}.

\subsection{Zero-shot Graph Classification}
\begin{table*}[t]
    \centering
    \caption{
       Zero-shot cross-domain graph classification results (Accuracy \% $\pm$ std and Macro-F1 \% $\pm$ std). The best results are highlighted in bold, and the second-best are underlined. Methods with "*" are directly cited from~\cite{TFSGFM}.
       }
    \label{tab:zero_shot_graph_classification}
    
    \renewcommand{\arraystretch}{1.15}
    \setlength{\tabcolsep}{7pt}
    
    \resizebox{\textwidth}{!}{
    \begin{tabular}{llcccccc}
        \toprule
        \multirow{2}{*}{\textbf{Setting}}
        & \multirow{2}{*}{\textbf{Method}}
        & \multicolumn{2}{c}{\textbf{IMDB-BINARY}}
        & \multicolumn{2}{c}{\textbf{ENZYMES}}
        & \multicolumn{2}{c}{\textbf{DD}} \\
        
        \cmidrule(lr){3-4}
        \cmidrule(lr){5-6}
        \cmidrule(lr){7-8}
        
        & & \textbf{Accuracy} & \textbf{Macro-F1}
          & \textbf{Accuracy} & \textbf{Macro-F1}
          & \textbf{Accuracy} & \textbf{Macro-F1} \\
        \midrule
        
        Supervised
        & GCN*
        & $57.30 \pm 0.98$
        & $54.62 \pm 1.12$
        & $20.58 \pm 2.00$
        & $15.25 \pm 3.96$
        & $44.74 \pm 4.23$
        & $55.33 \pm 6.22$ \\
        
        \midrule
        
        \multirow{3}{*}{One-shot}
        & GPPT*
        & $50.15 \pm 0.75$
        & $44.16 \pm 6.70$
        & $21.29 \pm 3.79$
        & $19.87 \pm 2.99$
        & $51.50 \pm 6.54$
        & $57.69 \pm 6.89$ \\
        
        & GPF*
        & $59.65 \pm 5.06$
        & $56.22 \pm 6.17$
        & $22.00 \pm 1.25$
        & $17.34 \pm 2.45$
        & $48.52 \pm 7.11$
        & $59.36 \pm 1.18$ \\
        
        & GPF-plus*
        & $57.93 \pm 1.62$
        & $55.55 \pm 2.03$
        & $22.92 \pm 1.64$
        & $18.39 \pm 2.76$
        & $46.24 \pm 4.86$
        & $57.62 \pm 2.42$ \\
        
        \midrule
        
        \multirow{4}{*}{Zero-shot}
        & ULTRA (3g)*
        & $49.25 \pm 0.00$
        & $38.87 \pm 0.00$
        & $15.21 \pm 0.00$
        & $5.84 \pm 0.00$
        & $63.50 \pm 0.00$
        & $70.38 \pm 0.00$ \\
        
        & SCORE*
        & $61.83 \pm 1.60$
        & $60.91 \pm 2.18$
        & $22.92 \pm 2.03$
        & $\bm{21.77 \pm 2.17}$
        & $69.85 \pm 0.51$
        & $69.96 \pm 0.74$ \\
        
        & TFSGFM*
        & \underline{$63.67 \pm 1.32$}
        & \underline{$63.25 \pm 1.37$}
        & \underline{$23.07 \pm 1.71$}
        & $19.22 \pm 1.83$
        & \underline{$73.54 \pm 0.78$}
        & \underline{$72.84 \pm 0.73$ }\\

        & \textbf{SliGFM(Ours)}
        & $\bm{64.56\pm1.00}$
        & $\bm{64.29\pm0.88}$
        & $\bm{23.96\pm1.45}$
        & \underline{$ 20.00\pm 1.75$}
        & $\bm{73.83 \pm 0.72}$
        & $\bm{73.14 \pm 0.64}$ \\
        
        \bottomrule
    \end{tabular}
    }
\end{table*}
Table~\ref{tab:zero_shot_graph_classification} reports the zero-shot cross-domain graph classification results. SliGFM achieves the best accuracy on all three datasets and the best Macro-F1 on IMDB-BINARY and DD, while remaining competitive on ENZYMES. Notably, without using labeled samples from the target datasets, SliGFM consistently surpasses the supervised and one-shot baselines in classification accuracy, further demonstrating its strong generalization capability under the more challenging zero-shot setting.

A notable result is that the node-level representations learned by SliGFM transfer effectively to whole graph classification. Without introducing a task-specific graph encoder or downstream adaptation, we obtain graph representations by simply averaging the pretrained node embeddings and achieve consistently strong performance on IMDB-BINARY, ENZYMES, and DD, despite their substantial differences in graph scale, connectivity, and feature distributions. This suggests that SliGFM captures complementary feature and structural information that remains discriminative after pooling. Topology-aware feature modeling retains transferable feature patterns, while structure-guided adaptive propagation incorporates neighborhood context at scales appropriate to each graph. Their combination allows simple mean pooling to produce expressive graph-level representations, further verifying the applicability of SliGFM across both different graph domains and different levels of downstream prediction.

\begin{figure}[!htp]
    \centering
    \includegraphics[width=1.0\linewidth]{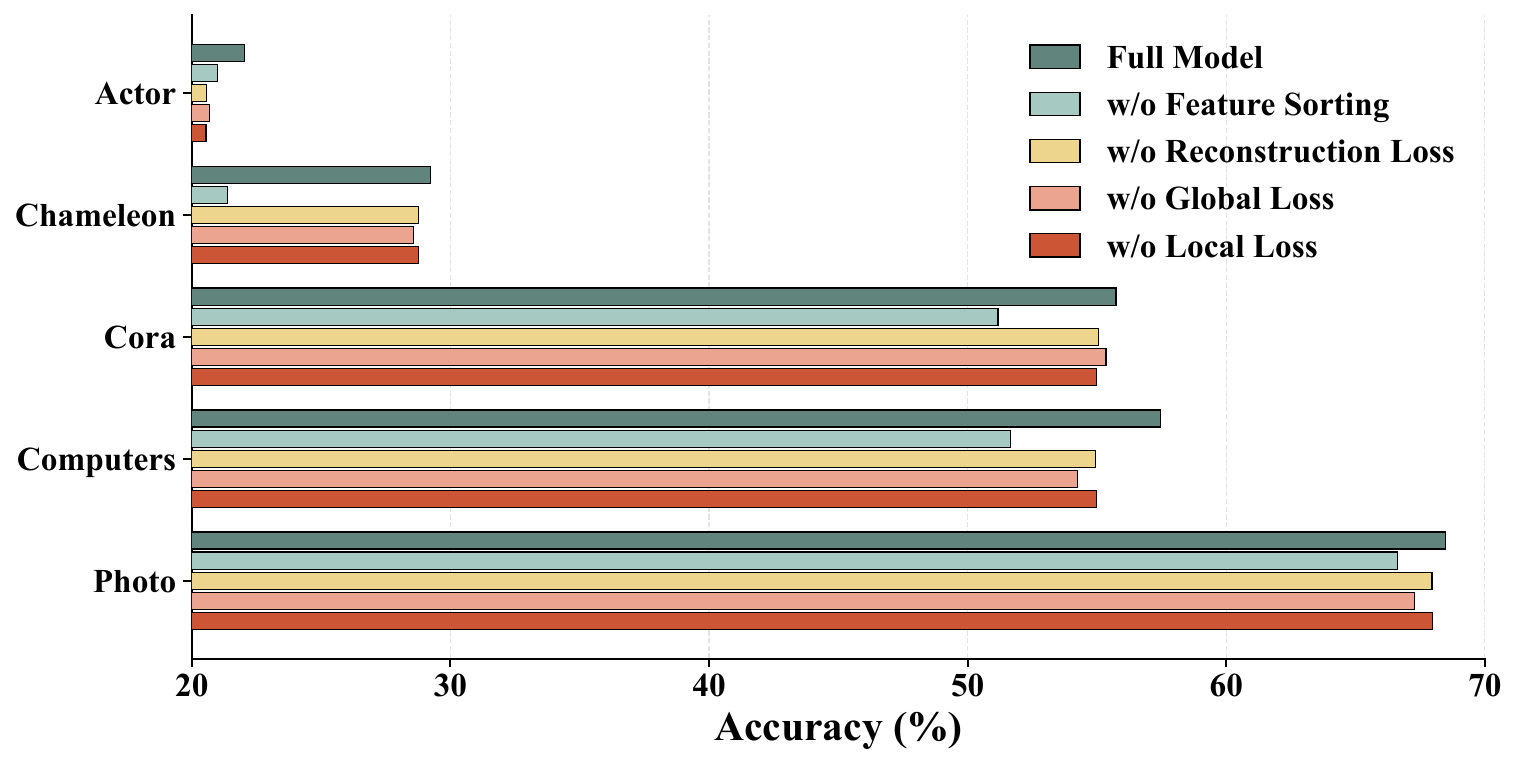}
    \caption{Ablation study on five cross-domain one-shot node classification datasets.}
    \label{fig:ablation}
\end{figure}

\subsection{Ablation Study}
To examine the contribution of each component in SliGFM, we construct four variants while keeping all other settings unchanged. Specifically, \textit{w/o Feature Sorting} retains the original feature order, while \textit{w/o Reconstruction Loss}, \textit{w/o Global Loss}, and \textit{w/o Local Loss} remove $\mathcal{L}_{\mathrm{recon}}$, $\mathcal{L}_{\mathrm{global}}$, and $\mathcal{L}_{\mathrm{local}}$, respectively.

As shown in Figure~\ref{fig:ablation}, the complete SliGFM consistently achieves the best performance across all five datasets, confirming the effectiveness of each design. Among the variants, removing feature sorting leads to the most pronounced degradation, especially on Chameleon, Cora, and Computers. This demonstrates that the original feature coordinates, which vary arbitrarily across datasets, are unfavorable for cross-domain transfer, whereas topology-aware ordering provides a more consistent basis for modeling heterogeneous attributes. Removing the reconstruction objective also causes consistent performance drops, indicating that preserving information from the original feature is important when learning unified representations. The global and local objectives provide further improvements by promoting discriminative representations at the global level while maintaining structural consistency among neighboring nodes. Overall, the results show that effective cross-domain representation learning requires not only a transferable organization of heterogeneous features, but also sufficient information preservation and complementary representation regularization.

\subsection{Hyperparameter Sensitivity}
We investigate the sensitivity of SliGFM to the sliding-window size $d_p$ in the topology-aware feature tokenizer. We vary $d_p$ over $\{64, 128, 256, 512\}$, while setting the sliding stride $\delta$ to one quarter of the corresponding window size, i.e., $\delta=d_p/4$, such that the overlap ratio between adjacent feature windows remains fixed.

\begin{figure}[!htp]
    \centering

    \begin{subfigure}[t]{0.48\columnwidth}
        \centering
        \includegraphics[width=\linewidth]{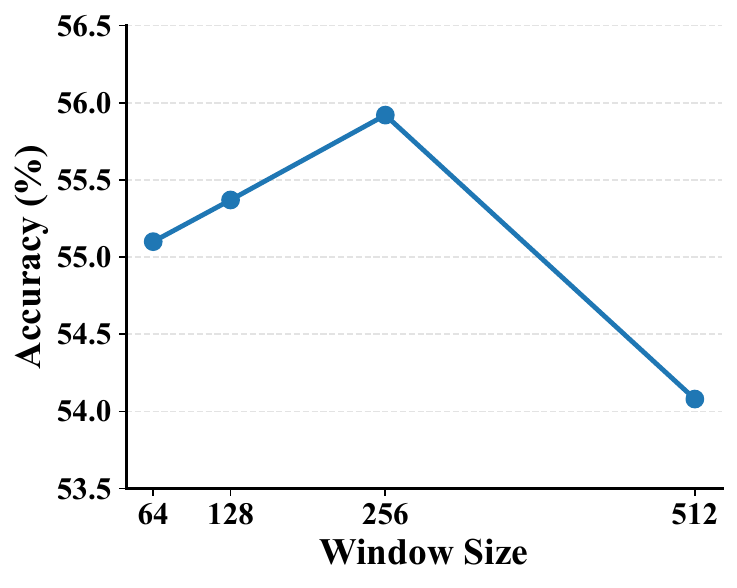}
        \caption{Cora}
        \label{fig:sub1}
    \end{subfigure}
    \hfill
    \begin{subfigure}[t]{0.48\columnwidth}
        \centering
        \includegraphics[width=\linewidth]{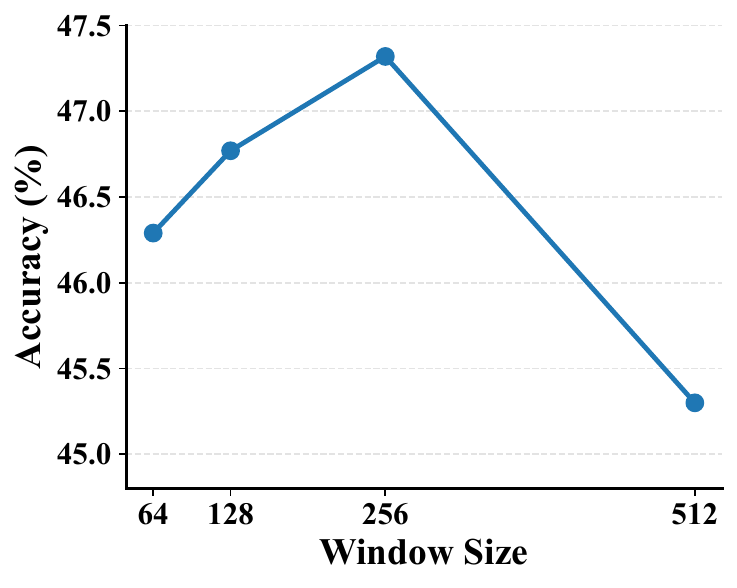}
        \caption{CiteSeer}
        \label{fig:sub2}
    \end{subfigure}

    \vspace{1mm}

    \begin{subfigure}[t]{0.48\columnwidth}
        \centering
        \includegraphics[width=\linewidth]{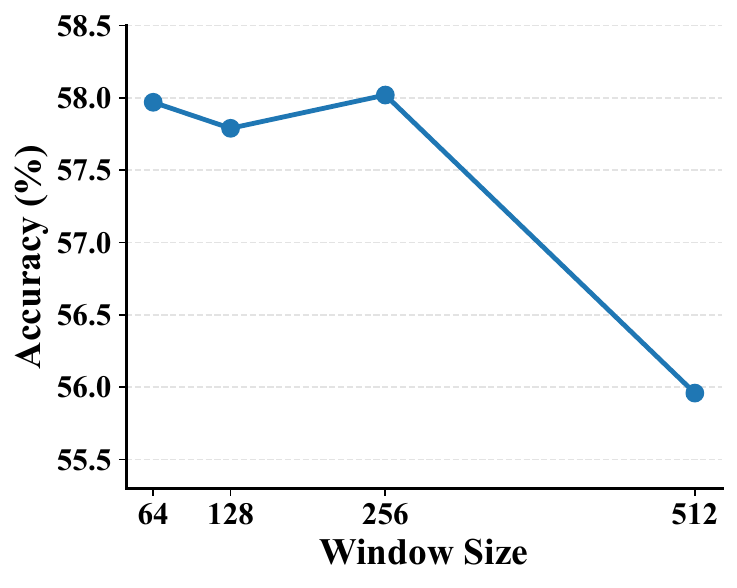}
        \caption{PubMed}
        \label{fig:sub3}
    \end{subfigure}
    \hfill
    \begin{subfigure}[t]{0.48\columnwidth}
        \centering
        \includegraphics[width=\linewidth]{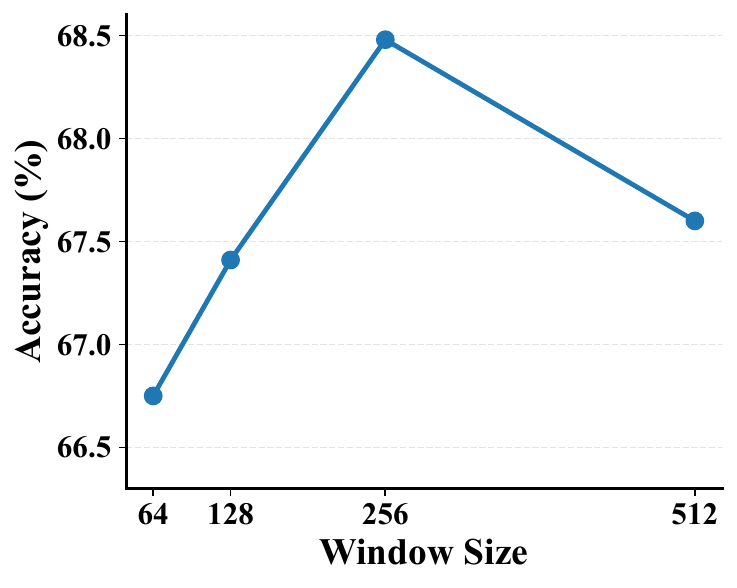}
        \caption{Photo}
        \label{fig:sub4}
    \end{subfigure}
    \caption{Parameter sensitivity analysis on the sliding window size $d_p$, with the sliding stride $\delta$ fixed at one quarter of the corresponding window size.}
    \label{fig:window_sensitivity}
\end{figure}

As shown in Figure ~\ref{fig:window_sensitivity}, SliGFM is relatively stable under moderate window sizes, with $d_p=256$ generally providing the best performance. A small window covers only a limited portion of the smoothness-ordered features, which may fragment correlated feature patterns and produce an unnecessarily long token sequence. Increasing the window size provides each token with richer local context and achieves a better balance between feature coverage and token granularity. However, when $d_p$ becomes excessively large, e.g., $512$, performance consistently declines across all four datasets. This is because overly large windows merge too many feature dimensions into a single token, leading to coarse-grained representations that may obscure informative local patterns. Overall, these results indicate that SliGFM is reasonably robust to moderate variations in $d_p$, while an appropriate window size is important for balancing local feature modeling and representation granularity.

\section{Conclusion}
In this work, we revisit cross-domain feature unification from the perspective of graph foundation models and argue that simply mapping heterogeneous node attributes into a common dimensional space is insufficient for effective cross-domain learning. We identify four desirable properties for unified graph features, including formal uniformity, cross-domain transferability, information preservation, and backbone compatibility. Based on these principles, we develop SliGFM, which establishes a topology-aware feature order and employs sliding-window feature encoding to transform heterogeneous node feature into unified token sequences. An intra-node Transformer further models dependencies among feature tokens, while structure-guided adaptive propagation incorporates target-specific graph topology into the learned node representations. Moreover, generative feature reconstruction encourages the unified representation to preserve the original information. Extensive experiments across heterogeneous graph datasets validate the effectiveness and generalization ability of SliGFM.

\bibliographystyle{ACM-Reference-Format}
\bibliography{references}

\appendix
\section{Notations and Definitions}
\label{appendix:notations}
The notations used in this paper and their descriptions are summarized in Table~\ref {tab:notations}.

\begin{table}[!htp]
\centering
\caption{Summary of the major notations.}
\label{tab:notations}
\fontsize{8.5pt}{12pt}\selectfont
\setlength{\tabcolsep}{1.5pt}
\renewcommand{\arraystretch}{0.9}

\begin{tabularx}{\columnwidth}{
    @{}
    >{\raggedright\arraybackslash}p{0.37\columnwidth}
    >{\raggedright\arraybackslash}X
    @{}
}
\hline
\textbf{Symbol} & \textbf{Description} \\
\hline

$\mathcal{G}$
& Attributed graph \\

$\mathcal{V},\mathcal{E}$
& Node and edge sets \\

$\mathbf{X},\mathbf{A}$
& Feature and adjacency matrices \\

$N,d$
& Node count and feature dimension \\

$\mathcal{N}(v_i)$
& One-hop neighbors of $v_i$ \\

$\mathcal{D}_{\mathcal{G}},M$
& Graph collection and its size \\

$\mathcal{G}^{(m)},N_m,d_m$
& Graph $m$, its size and dimension \\

$\mathcal{U},\bar{\mathcal{X}}$
& Feature unifier and shared space \\

\hline

$s_j^{(m)}$
& Feature smoothness score \\

$\boldsymbol{\pi}^{(m)}$
& Topology-aware feature order \\

$\bar{\mathbf{X}}^{(m)}$
& Reordered feature matrix \\

$d_p,\delta$
& Patch size and stride \\

$\mathbf{p}_{i,k}^{(m)}$
& Patch $k$ of node $v_i^{(m)}$ \\

$E_{\theta},d_t$
& Patch encoder and token dimension \\

$\mathbf{t}_{i,k}^{(m)}$
& Token of patch $k$ \\

$\mathbf{T}_i^{(m)},K_m$
& Token sequence and its length \\

\hline

$\mathbf{c}_i^{(m)}$
& Initial node-specific \textit{[CLS]} token \\

$\mathbf{U}_i^{(m,l)}$
& Token sequence at layer $l$ \\

$\bar{s}_k^{(m)},\mathbf{B}^{(m)}$
& Token smoothness and bias matrix \\

$\mathbf{z}_{i,\mathrm{cls}}^{(m)}$
& Contextualized \textit{[CLS]} embedding \\

$\widehat{\mathbf{T}}_i^{(m)}$
& Contextualized feature tokens \\

\hline

$\mathbf{g}^{(m)}$
& Graph structural statistics \\

$\boldsymbol{\alpha}^{(m)}$
& Graph-specific hop weights \\

$\widehat{\mathbf{A}}^{(m)}$
& Normalized adjacency matrix \\

$K_{\mathrm{hop}}$
& Maximum propagation hops \\

$\mathbf{Z}^{(m)}$
& Final node embeddings \\

\hline

$\mathbf{R}_{\mathrm{qry}}^{(K_m)}$
& Learnable reconstruction queries \\

$\operatorname{Decoder}_{\omega}$
& Token reconstruction decoder \\

$D_{\xi},\widehat{\mathbf{p}}_{i,k}^{(m)}$
& Patch decoder and reconstruction \\

$\mathbf{z}_i^{(m)},\bar{\mathbf{z}}^{(m)}$
& Node embedding and graph mean \\

$\mathcal{L}_{\mathrm{recon}}$
& Reconstruction loss \\

$\mathcal{L}_{\mathrm{local}}$
& Local alignment loss \\

$\mathcal{L}_{\mathrm{global}}$
& Global dispersion loss \\

$\mathcal{L}$
& Total pretraining loss \\

\hline
\end{tabularx}
\end{table}

\section{Detailed Dataset Descriptions}
\label{appendix:datasets}

\begin{table*}[!htp]
\centering
\caption{Statistics of node and graph classification datasets.}
\label{tab:dataset_statistics}
\resizebox{\textwidth}{!}{
\begin{tabular}{llllllll}
\toprule
\textbf{Category} & \textbf{Dataset} & \textbf{\#Graphs} & \textbf{\#Avg.nodes} & \textbf{\#Avg.edges} & \textbf{\#Features} & \textbf{\#Classes} & \textbf{Domain} \\
\midrule

\multirow{6}{*}{\makecell[l]{Homophilic \\ node classification}}
& Cora      & 1 & 2,708  & 10,556  & 1,433 & 7  & Citation networks \\
& CiteSeer  & 1 & 3,327  & 9,104   & 3,703 & 6  & Citation networks \\
& PubMed    & 1 & 19,717 & 88,648  & 500   & 3  & Citation networks \\
& Computers & 1 & 13,752 & 491,722 & 767   & 10 & Co-purchase networks \\
& Photo     & 1 & 7,650  & 238,162 & 745   & 8  & Co-purchase networks \\
& CS        & 1 & 18,333 &163788  & 6805   & 15  & Co-author networks \\
\midrule

\multirow{3}{*}{\makecell[l]{Heterophilic \\ node classification}}
& Squirrel  & 1 & 5,201 & 217,073 & 2,089 & 5 & Wikipedia networks \\
& Chameleon & 1 & 2,277 & 36,101  & 2,325 & 5 & Wikipedia networks \\
& Actor   & 1 &  7600  & 30,019    & 932 & 5  & Co-occurrence networks\\
\midrule

\multirow{3}{*}{\makecell[l]{Graph \\ classification}}
& IMDB-BINARY & 1,000 & 19.8  & 212.8   & 136 & 2 & Social networks \\
& ENZYMES     & 600   & 32.6  & 156.9   & 3   & 6 & Biological networks \\
& DD          & 1,178 & 284.3 & 1,715.6 & 89  & 2 & Biological networks \\
\bottomrule
\end{tabular}
}
\end{table*}

For homophilic node classification, we use Cora, CiteSeer, and PubMed, which are citation networks where nodes represent papers and edges represent citation relationships; node features are bag-of-words representations of paper content, and labels correspond to the research topic of each paper. Computers and Photo are co-purchase networks extracted from Amazon, where nodes represent goods and edges indicate that two goods are frequently purchased together; node features are bag-of-words representations of product reviews, and labels denote the product category. CS is a co-authorship network extracted from the Microsoft Academic Graph, where nodes represent authors and edges indicate co-authorship relationships; node features encode keywords of each author's publications, and labels denote the author's most active research field. For heterophilic node classification, we use Squirrel and Chameleon, which are Wikipedia networks where nodes represent web pages on specific topics and edges represent mutual hyperlinks between pages; node features correspond to informative nouns extracted from the pages, and the average monthly traffic of each page determines labels. Actor is a co-occurrence network extracted from film industry data, where nodes represent actors and edges indicate co-occurrence on the same Wikipedia page; node features are derived from keywords in the actors' Wikipedia pages, and labels are determined accordingly. For graph classification, we use IMDB-BINARY, a movie collaboration dataset where each graph corresponds to an ego-network of actors who appear together in a movie, with graph labels indicating the movie genre. ENZYMES is a dataset of protein tertiary structures, in which nodes represent secondary structure elements, and edges represent their spatial or sequential neighboring relationships, with each graph labeled according to one of six enzyme classes. DD is a dataset of protein structures, where nodes represent amino acids, and edges connect two amino acids if they are less than 6 Angstroms apart, with graphs labeled as enzymes or non-enzymes. Detailed statistics of all datasets are summarized in Table~\ref{tab:dataset_statistics}.

\section{Detailed Experiment Settings}
\label{appendix:experiment}
For few-shot cross-domain node classification, we jointly pre-train SliGFM on Cora, PubMed, Computers, and Chameleon. During downstream evaluation, all pretrained parameters are kept frozen. For each target dataset, a small number of labeled nodes are sampled from each class as support examples, whose embeddings are averaged to construct the corresponding class prototypes. Each query node is then assigned to the class whose prototype is most similar to its embedding. 
For zero-shot graph classification, SliGFM is pretrained at the graph level. During downstream evaluation, the frozen model first produces node embeddings for each subgraph, which are subsequently aggregated through average pooling to obtain a graph-level representation. Following the evaluation protocol of TFSGFM~\cite{TFSGFM}, class prototypes are constructed by averaging the representations of labeled training graphs within each class. Each test graph is then classified according to the similarity between its representation and the corresponding class prototypes. All experiments are conducted on a server equipped with an Intel Xeon Gold 5218R CPU and an NVIDIA GeForce RTX 3090 GPU.

\begin{table*}[!htp]
\centering
\caption{Three-shot cross-domain node classification results (Accuracy \% $\pm$ std). The best results are highlighted in bold, and the second-best are underlined.}
\label{tab:node_3shot}
\resizebox{\textwidth}{!}{
\begin{tabular}{lccccccccc}
\toprule
\textbf{Method}
& \textbf{Cora}
& \textbf{CiteSeer}
& \textbf{PubMed}
& \textbf{Computers}
& \textbf{Photo}
& \textbf{CS}
& \textbf{Squirrel}
& \textbf{Chameleon}
& \textbf{Actor}\\
\midrule

MDGFM
& $51.80\pm4.84$
& $48.83\pm5.50$
& $57.94\pm5.15$
& $66.38\pm9.02$
& $\underline{73.41\pm5.16}$
& $82.80\pm2.53$
& $22.49\pm2.69$
& $25.27\pm4.04$
& $20.16\pm1.38$
\\

MDGPT
& $54.91\pm5.56$
& $55.04\pm6.06$
& $61.21\pm7.73$
& $65.32\pm6.49$
& $72.42\pm7.00$
& $79.38\pm4.37$
& $23.14\pm3.12$
& $25.39\pm4.16$
& $20.31\pm1.76$
\\

BRIDGE
& $55.58\pm6.70$
& $\underline{56.85\pm6.74}$
& $61.31\pm5.53$
& $\underline{66.92\pm6.93}$
& $71.28\pm7.42$
& $80.24\pm4.57$
& $\underline{23.35\pm3.96}$
& $25.40\pm3.52$
& $19.92\pm1.39$
 \\

FUG
& $63.89\pm6.25$
& $49.89\pm6.30$
& $61.90\pm6.45$
& $61.11\pm6.40$
& $72.82\pm6.36$
& $85.20\pm3.03$
& $20.91\pm1.03$
& $24.20\pm1.62$
& $20.79\pm2.09$
\\

LEDA
& $58.19\pm7.18$
& $60.45\pm4.45$
& $\underline{64.36\pm7.36}$
& $60.26\pm10.11$
& $66.45\pm8.68$
& $80.74\pm1.76$
& $20.95\pm1.90$
& $\underline{26.93\pm4.75}$
& $21.45\pm2.94$
\\

SAMGPT
& $59.42\pm4.78$
& $54.58\pm7.01$
& $54.81\pm6.28$
& $59.91\pm6.68$
& $73.14\pm5.26$
& $84.04\pm2.28$
& $\bm{23.43\pm3.42}$
& $25.15\pm3.71$
& $20.09\pm1.48$
\\

TIG
& $63.81\pm6.62$
& $54.26\pm5.64$
& $54.67\pm8.31$
& $60.71\pm7.94$
& $64.82\pm8.43$
& $83.61\pm1.84$
& $20.33\pm0.71$
& $24.20\pm4.41$
& $20.15\pm2.29$
\\

TFSGFM
& $\underline{64.88\pm5.56}$
& $\bm{57.76\pm5.57}$
& $60.96\pm6.13$
& $62.51\pm10.81$
& $73.02\pm7.76$
& $\underline{86.59\pm1.79}$
& $20.28\pm1.52$
& $26.71\pm3.30$
& $\underline{25.05\pm3.14}$
\\

\midrule

\textbf{Ours}
& $\bm{65.11\pm5.47}$
& $\bm{57.76\pm5.46}$
& $\bm{65.56\pm7.16}$
& $\bm{67.98\pm7.24}$
& $\bm{74.90\pm6.32}$
& $\bm{87.28\pm1.83}$
& $21.34\pm2.00$
& $\bm{30.22\pm3.26}$
& $\bm{25.14\pm3.65}$\\

\bottomrule
\end{tabular}
}
\end{table*}

\section{Supplementary Experiment Results}
\label{appendix:few-shot}
We present the detailed Accuracy of SliGFM under the 3-shot and 5-shot settings in Table~\ref{tab:node_3shot} and ~\ref{tab:node_5shot}.

\begin{table*}[!htp]
\centering
\caption{Five-shot cross-domain node classification results (Accuracy \% $\pm$ std). The best results are highlighted in bold, and the second-best are underlined.}
\label{tab:node_5shot}
\resizebox{\textwidth}{!}{
\begin{tabular}{lccccccccc}
\toprule
\textbf{Method}
& \textbf{Cora}
& \textbf{CiteSeer}
& \textbf{PubMed}
& \textbf{Computers}
& \textbf{Photo}
& \textbf{CS}
& \textbf{Squirrel}
& \textbf{Chameleon}
& \textbf{Actor} \\
\midrule

MDGFM
& $59.54\pm3.28$
& $52.49\pm3.46$
& $62.94\pm5.55$
& $\underline{69.88\pm6.49}$
& $79.35\pm3.60$
& $86.14\pm1.62$
& $22.25\pm2.73$
& $26.72\pm3.81$
& $20.72\pm1.35$ \\

MDGPT
& $58.91\pm4.26$
& $58.86\pm3.46$
& $64.71\pm8.19$
& $68.48\pm4.89$
& $79.59\pm4.15$
& $81.39\pm3.47$
& $21.92\pm4.02$
& $25.34\pm3.91$
& $20.73\pm1.66$ \\

BRIDGE
& $61.06\pm4.41$
& $60.11\pm3.87$
& $63.69\pm6.21$
& $69.57\pm5.94$
& $74.48\pm6.99$
& $82.16\pm3.59$
& $\underline{22.40\pm4.05}$
& $25.37\pm3.72$
& $20.58\pm1.39$ \\

FUG
& $68.56\pm5.27$
& $53.46\pm5.54$
& $66.84\pm6.31$
& $66.15\pm6.03$
& $78.75\pm4.92$
& $87.39\pm2.33$
& $20.88\pm1.24$
& $24.63\pm2.21$
& $20.46\pm1.73$ \\

LEDA
& $63.47\pm6.10$
& $\bm{63.52\pm2.23}$
& $\underline{67.49\pm6.39}$
& $59.58\pm10.71$
& $66.31\pm7.92$
& $82.08\pm1.99$
& $20.21\pm1.71$
& $27.52\pm4.09$
& $21.68\pm3.27$ \\

SAMGPT
& $64.68\pm3.50$
& $60.99\pm4.02$
& $59.14\pm7.68$
& $63.78\pm5.93$
& $77.29\pm3.53$
& $86.02\pm1.84$
& $\bm{22.78\pm3.32}$
& $25.30\pm3.71$
& $20.36\pm1.37$ \\

TIG
& $67.50\pm4.28$
& $58.56\pm4.87$
& $55.06\pm7.47$
& $63.10\pm4.79$
& $67.68\pm7.04$
& $85.27\pm1.24$
& $20.16\pm0.59$
& $24.45\pm4.82$
& $20.63\pm1.94$ \\

TFSGFM
& $\underline{68.64\pm4.30}$
& $61.84\pm3.58$
& $63.78\pm5.27$
& $67.71\pm7.68$
& $72.73\pm7.40$
& $\underline{88.28\pm1.10}$
& $20.44\pm1.53$
& $\underline{27.80\pm2.82}$
& $\underline{25.68\pm3.09}$ \\

\midrule

\textbf{Ours}
& $\bm{68.85\pm3.46}$
& $\underline{62.60\pm3.90}$
& $\bm{67.77\pm5.47}$
& $\bm{71.74\pm4.79}$
& $73.22\pm6.34$
& $\bm{88.58\pm0.95}$
& $21.64\pm1.94$
& $\bm{30.62\pm2.57}$
& $\bm{26.34\pm3.08}$ \\

\bottomrule
\end{tabular}
}
\end{table*}
\end{document}